\documentclass{llncs}
\usepackage{llncsdoc}
\usepackage{graphicx}
\usepackage[T1]{fontenc}
\usepackage[utf8]{inputenc}
\usepackage{authblk}
\usepackage{amsmath}
\usepackage{graphicx}
\usepackage{caption}
\usepackage{subcaption}
\usepackage{lipsum}
\usepackage{dblfloatfix}
\usepackage{url}
\usepackage[usenames, dvipsnames]{color}

\begin{document}

\title{Ensemble Sales Forecasting Study in Semiconductor Industry} 
\author{Qiuping Xu\inst{1} \and Vikas Sharma\inst{2}}
\institute{Intel Corporation. \email{\{qiuping.xu, vikas.sharma\}@intel.com}
}

\maketitle


\begin{abstract} 
Sales forecasting plays a prominent role in business planning and business strategy. The value and importance of advance information is a cornerstone of planning activity, and a well-set forecast goal can guide sale-force more efficiently. A forecasting usually depends on many factors such as the product feature, supply chain constrain, market demand, market share, promotion strategy, competition, macroeconomics condition and others. However, most of those data is hard or even impossible to collect. In this paper CPU sales forecasting of Intel Corporation, a multinational semiconductor industry, was considered. We consolidated the available data resource and forecasting requirement, matched them against the optimal methodology. Past sale, future booking, exchange rates, Gross domestic product (GDP) forecasting, seasonality and other indicators were innovatively incorporated into the quantitative modeling. Benefit from the recent advances in computation power and software development, millions of models built upon multiple regressions, time series analysis, random forest and boosting tree were executed in parallel. The models with smaller validation errors were selected to form the ensemble model. To better capture the distinct characteristics, forecasting models were implemented at lead time and lines of business level. The moving windows validation process automatically selected the models which closely represent current market condition. The weekly cadence forecasting schema allowed the model to response effectively to market fluctuation. Generic variable importance analysis was also developed to increase the model interpretability. Rather than assuming fixed distribution, this non-parametric permutation variable importance analysis provided a general framework across methods to evaluate the variable importance. This variable importance framework can further extend to classification problem by modifying the mean absolute percentage error(MAPE) into misclassify error. This forecast output now helps formulate part of the input provided to public and investors as guideline for the following quarter during Intel's quarterly earning release\footnote{Please find the demo code at \\\url{https://github.com/qx0731/ensemble_forecast_methods}.}. 

\end{abstract}

\section{Background and Motivation}
\label{sec:background}
Sales forecasting is the foundation for planning various phases of the company's business operations. 
A sales forecast predicts the value of sales over a period of time. Marketing and other managerial functions need different types of forecasting horizons because each directly affects a different business function. The work presented in this paper was focusing on short term forecast. The forecast was made for tactical reasons that included production planning, sale target setup, short-term cash requirements and adjustments that needed to be made for sales fluctuations.

For the sale forecasting to be accurate, all (not limited to) of the following factors need to be considered: historical perspective, economic conditions, expected market share, total available market, manufacturing constrains, efficiency of distribution channel. The sensitivity of those parameters also needs to be considered and incorporated. However, due to the complexity of the markets and low visibility of data resources, alternative data needs to be considered and methodology needs to be developed to better describe market's dynamic. Thus, the sale forecasting effort is still an active field that attracts extensive attention from both industry and academia.

Research in sales forecasting can be traced back to 1950s \cite{Boulden1958}. Since then a large number of sales forecasting papers have been published, in which various forecasting techniques have been proposed. The most commonly used techniques for sales forecasting include statistically based techniques like time series \cite{Winters1960,Box1969,Groff1973} regression techniques \cite{Jain2007}, artificial neural networks \cite{Giles2001,Huang2004,Sharma2012} and other hybrid models \cite{Clements2002,Ferreira2015}.

No surprise that each type of the models has its particular advantages and disadvantages compared to other approaches. In this paper, we investigated and compared the performance of different techniques on Intel's weekly CPU sale forecast. In particular, Extreme Gradient Boosting algorithm, Random forest, ensembled linear regression and ensembled autotegrreive integrated moving average models were considered. Given the fact that neural network performs generally well for sales forecasting if the demand is not seasonal and quite non fluctuating \cite{Wong2010}, and taking interpretability and data size into account, we left the direct implementation of neural network out the discussion of this paper. The comprehensive comparisons among those methods were illustrated across product segments and lead times. 
Undoubtedly, this research will not only provide an accurate basis of demand forecast, but also greatly rich the study on selecting and benchmark forecasting techniques for practical applications .

The rest of this paper is organized as follows: Sec.\ref{sec:data} includes the data description of this work; Sec.\ref{sec:method} describes forecasting models, moving windows validation process and variable importance extraction methodology; Empirical results and insights are included in Sec.\ref{sec:result}; Finally, Sec.\ref{sec:conclusion} concludes the paper with a summary of our results and potential areas for future work.

\section{Data and Feature Engineering}
\label{sec:data}
\begin{figure}[ht]
	\centering 
    \includegraphics[width=0.75\textwidth]{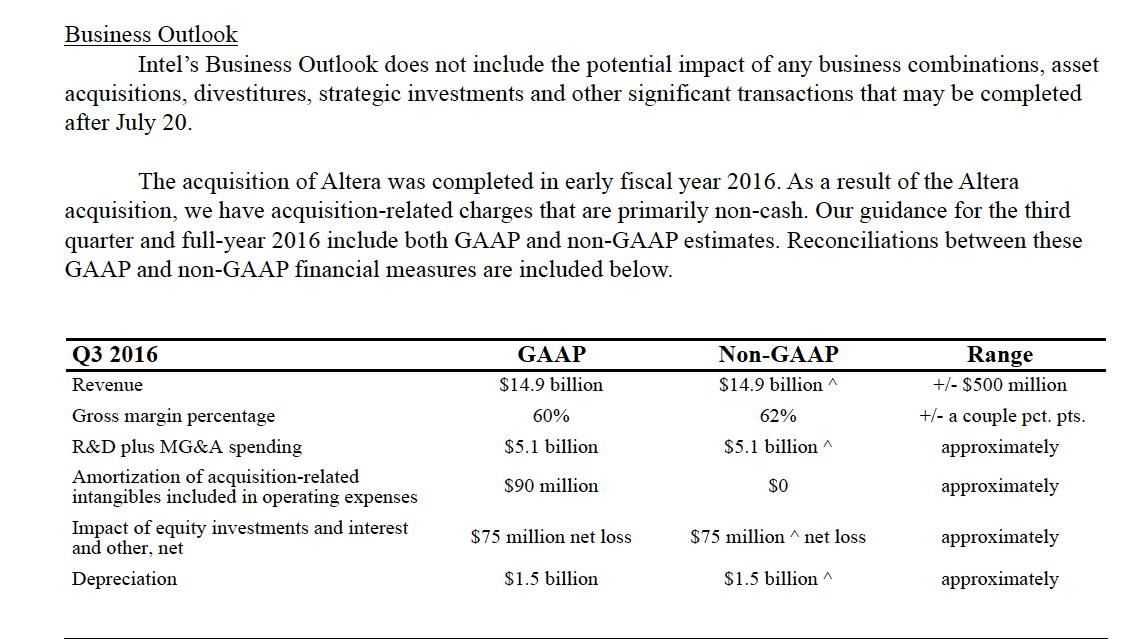}
    \caption{Part of outlook for Q3'16 from Intel's Q2'16 earning report}
    \label{fig:variable1}
\end{figure}

As the leading company in Semiconductor industry, Intel combines advanced chip design capability with a leading-edge manufacturing capability and be able to satisfy customers' needs along different product lines. By considering maturity levels of business segments, the CPU prediction problem was further divided into three sub problems - Desktop CPU, Notebook CPU and Server CPU sale forecast. One usage of this model was to be incorporated into the forecast pipeline to help generate the outlook for the next quarter as included in Intel's earning report Fig.\ref{fig:variable1} to guide company's execution. 

In this project, we were provided weekly sales data from 2012. This data represented a time-stamped weekly total sale of Intel's CPU by line of business. In addition, we were provided the historical back-then booking data, which showed a screen shot of booking information at a time in the past. Historical average selling price was also provided.

Besides internal data, we also derived variables from the following data.
\begin{itemize}
\item Foreign exchange outlook (back-then and forward look quarterly forecast): the forecast exchange rate between RMB(Currency of China) and US dollar, EURO and US dollar. 
\item GDP outlook (back-then and forward look quarterly forecast): GDP quarterly YoY\footnote{The rate of change between current quarter and the same quarter of previous year} forecast at world wide level and some key regions (Mainland China and Europe).
\item Seasonality: varies from 1 to 4 to indicate the four seasons of a year.
\item WeekofQuarter: varies from 1 to 13. This variable was used to indicate in quarter week. 
\item Special events: Chinese New Year (CNY), indicates the effect of CNY (0-1). We spread the CNY effect evenly to the +/-5 days around Lunar new year, and the summarized effect was computed weekly. 
\item Time-stamp: the quarters in this dataset were assigned to a list of consecutive numbers start from 1. This variable was used to capture the general growth/shrink of the business. 
\end{itemize}

Measuring GDP requires adding up the value of what is produced, net of inputs, across a wide variety of business lines, weighting each according to its importance in the economy. Thus, GDP statistics are so prone to constant and substantial measurement errors and subjective bias. Similar problem exists in foreign exchange outlooks. To overcome this difficulty, instead of taking the absolute numbers from the outlook, we construed variables as the differences among consecutive outlooks to reduce the subjective bias and measurement errors. The negative change in consecutive GDP outlooks indicated a loss of confidence in the quarter of interest. 

\begin{figure}[ht]
	\centering 
    \includegraphics[width=0.45\textwidth]{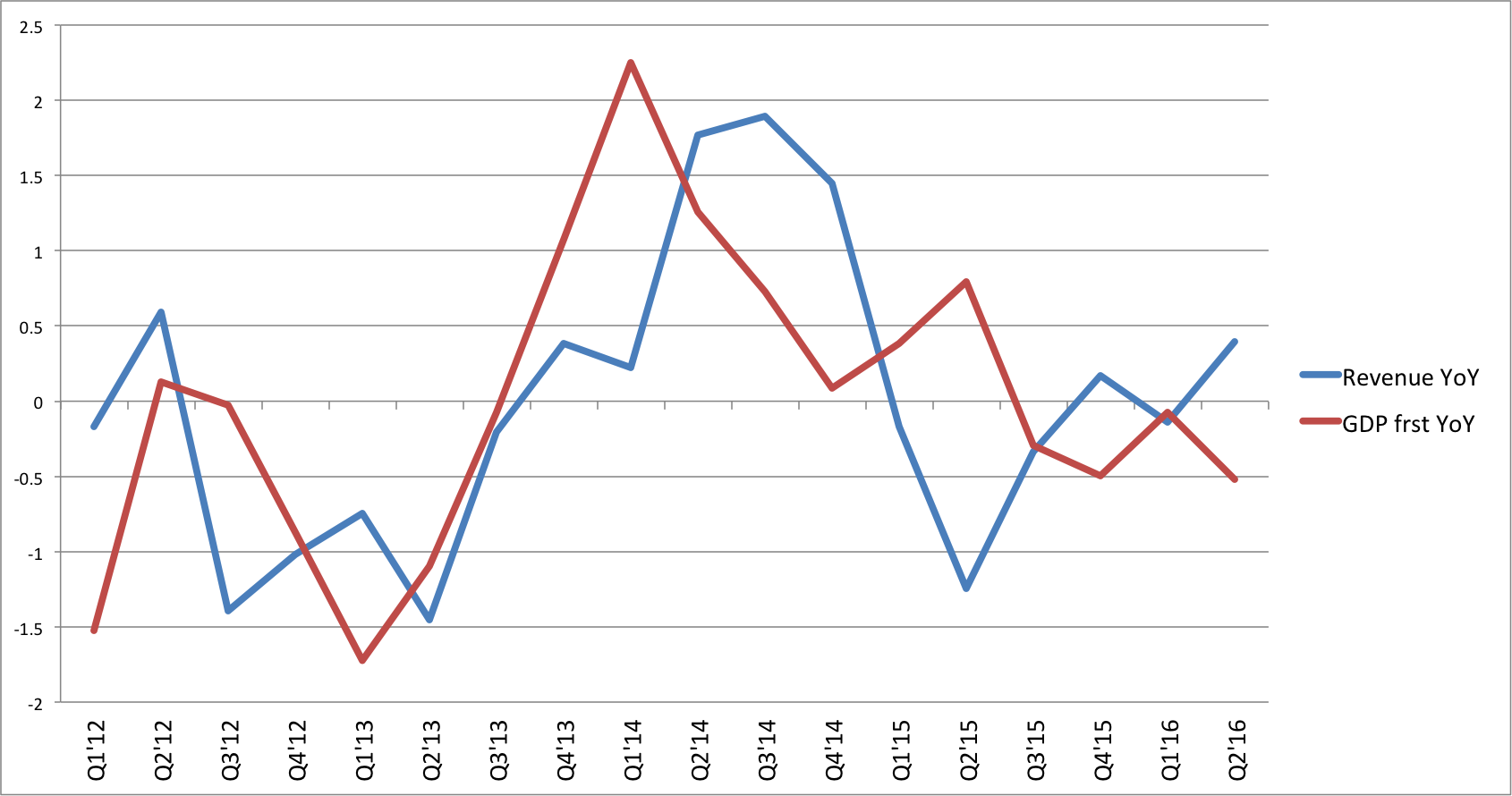}
    \caption{World wide GDP YoY (Red) with Intel revenue YoY (Blue) from Q1'12 to Q2'16.}
    \label{fig:variable2}
\end{figure}

Figure \ref{fig:variable2} showed the world wide GDP YoY with the Intel revenue YoY \footnote{14 weeks Q1'16 has been normalize to 13 weeks by a factor of $\frac{13}{14}$} for same period of time (Q1'12 to Q2'16). In order to visualize those two series together, those two series were normalized to have mean of zero and standard deviation of one. This picture revealed the fact that GDP plays an important role in revenue forecasting, especially for the points of inflection.

\section{Methods and Technical Solutions}
\label{sec:method}
In this section, we started with our approach for dealing with multicollinearity. The forecast methodologies were briefly discussed afterwards. To incorporate the nature of time dependence in the data, a dynamic moving windows validation method was proposed. A general framework of variable importance was also implemented to better interpret the model output.

\subsection{Multicollinearity}
\label{sec:colinear}
Multicollinearity is a phenomenon in which two or more predictor variables in a multiple regression model are highly correlated. Multicollinearity generally does not reduce the predictive power, it only affects calculations regarding individual predictors. In order to deal with this problem without giving up the interpretability, we took the approach of grouping similar numerical variables together and selecting the most representative variable from each group into the predictive model.

\begin{figure}[ht]
\begin{subfigure}{0.45\textwidth}
    \includegraphics[width=1\textwidth]{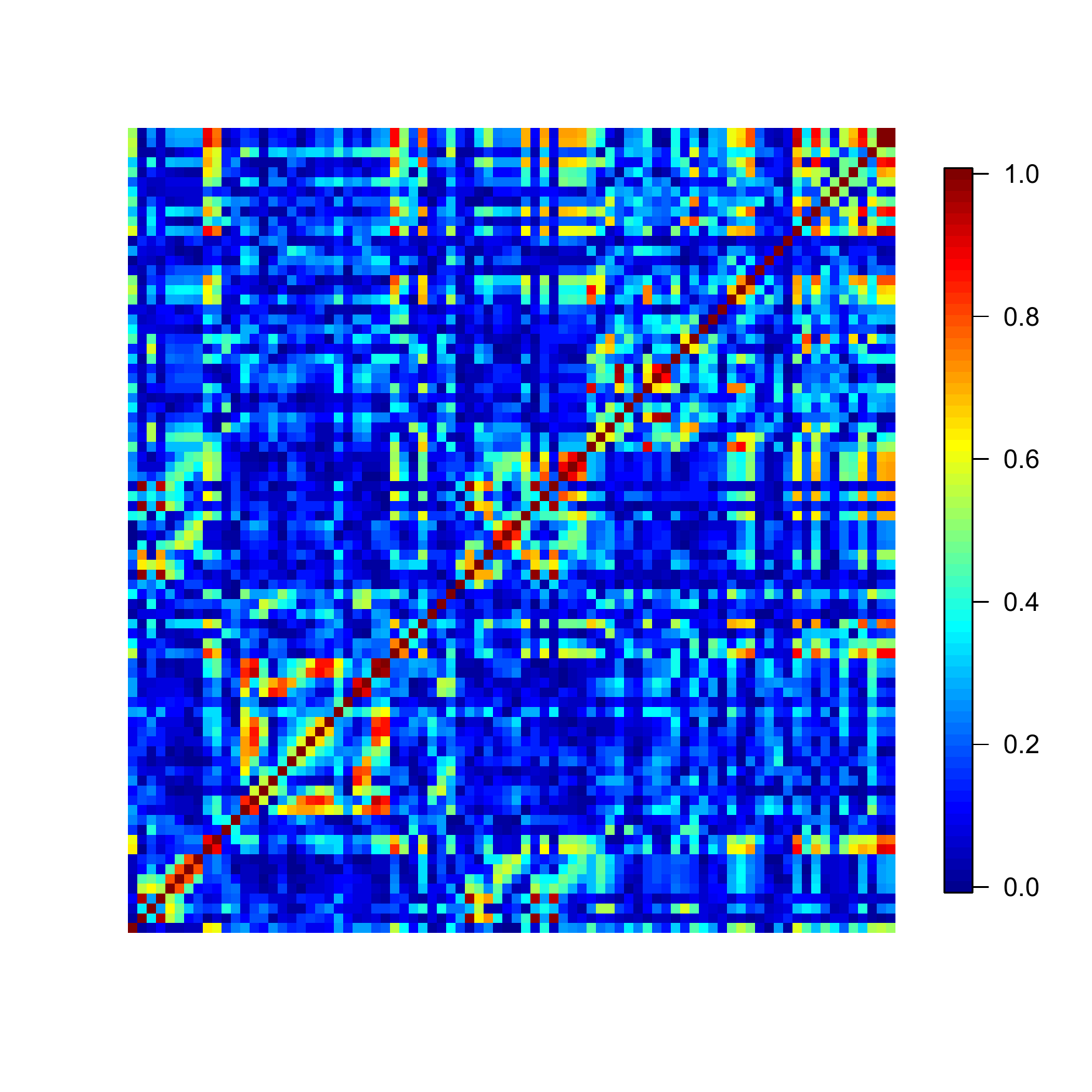}
    \caption{}
    \label{fig:before}
\end{subfigure}
~
\begin{subfigure}{0.45\textwidth}
    \includegraphics[width=1\textwidth]{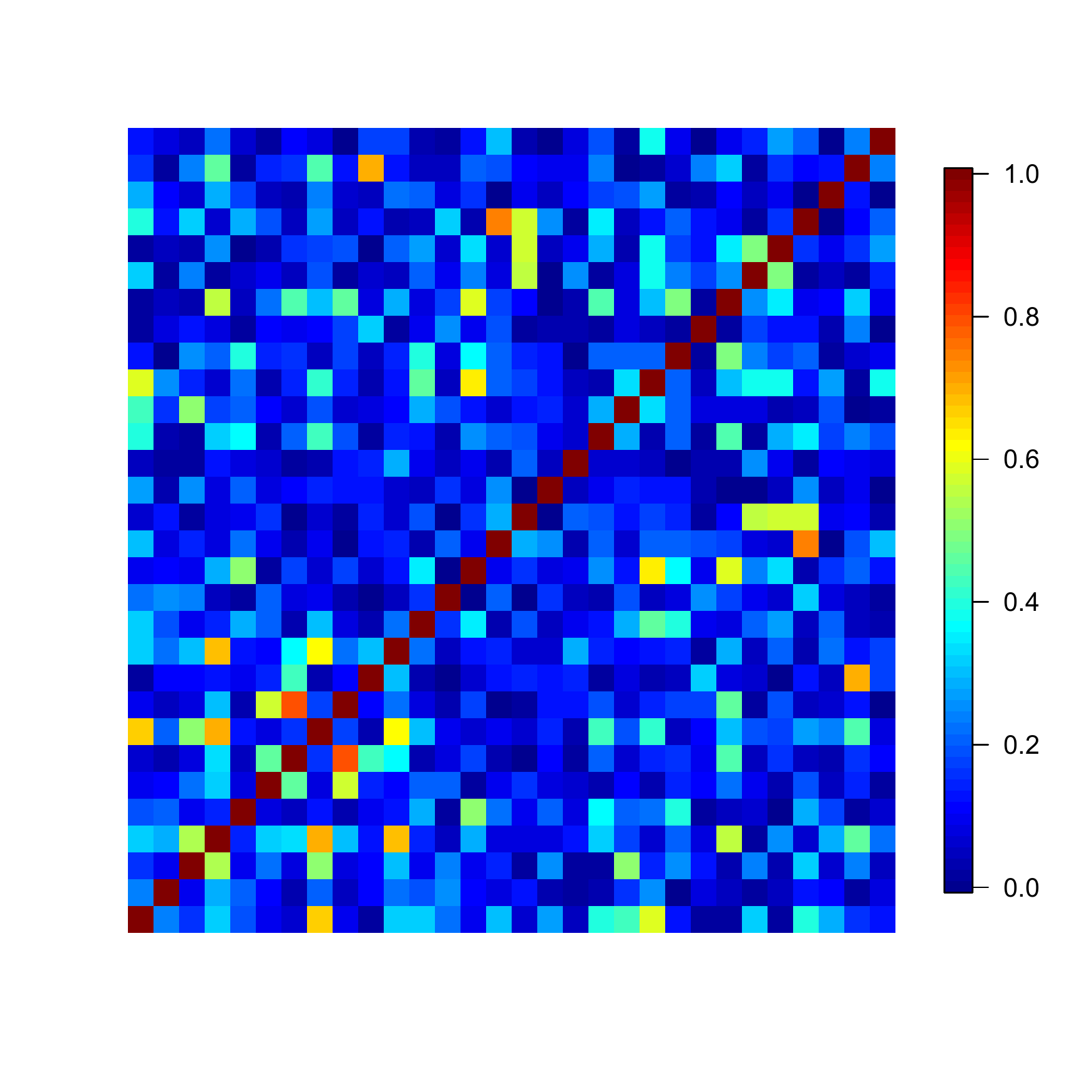}
    \caption{}
    \label{fig:after}
\end{subfigure}
\caption{The heat map of absolute correlation coefficients between pairs of variables before (a) and after(b) multicolinearity treatment.}
    \label{fig:colinear}
\end{figure}

The similarity metric was defined as the absolute correlation coefficients between pair of variables. To cluster the variables, multiple dimensional scaling (MDS) \cite{Kruskal1978} and K-means \cite{MacQueen1967} were used. A pair-wise distance matrix among the variables was construed as 1-the pair-wise absolute correlation coefficients matrix. MDS algorithm was then applied to find the best representation in $N$-dimensional space such that the pair-variable distances are preserved as well as possible. The $N$ in our experiment was selected to be the maximal to preserve as much information as possible. Then the mapping of MDS became the input of the K-means algorithm to form clusters. The number of clusters was determined by the ratio between  `between cluster variance' and  `total variance' (in the experiments, we set this number to be 80\%). Within each cluster, the variable had the maximal absolute correlation coefficient (with response variable) was selected into the modeling steps. One example of before and after the multicolinearity treatment was shown in Fig. \ref{fig:colinear}. In this example, the variance inflation factor was reduced from 166 to 4.3.

\subsection{Extreme Gradient Boosting}
XGBoost \cite{Chen2016} is short for Extreme Gradient Boosting. This is an efficient and scalable implementation of gradient boosting framework proposed by Friedman \cite{Friedman2001}. XGBoost constructs an additive model while optimizing a loss function. The loss functions not only accounts for the inaccuracy of the prediction (size of the residuals) but also considers a regularization term that controls the complexity of the model. The regularization helps to avoid over-fitting and has shown good performance in various Kaggle challenges. Empirical evidence shows that the boosting approach with small learning rate yields more accurate results\cite{Chen2016}. In the current project, we used $0.01$ as learning rate and $1000$ as the number of trees for one set of boosting experiment.

\subsection{Random Forest}
Random forest (RF), which was first proposed by Breiman \cite{Breiman2001}, is an ensemble of $B$ trees $f_{1}(X), \cdots, f_{B}(X)$, where $X = X_{1}; \cdots; X_{n}$ is variable matrix of $n$ observations, and $X_{i}$ is the $p$-dimension feature vector of each observation. $B$ is a predefined hyperparameter in RF algorithm. Unlike boosting tree, in random forest each tree is built on a bootstrap data set, independent of the other trees. RF also benefits from random feature selection to decorrelating the trees, thereby making the average of the resulting trees less variable and hence more reliable \cite{James2013}. This `small' tweak provides RF a large improvement over pure bagged trees.

\subsection{Parametric Methods}
In this section, we would like to briefly sketch two long standing methods in forecasting problems: multiple linear regression(MLR) and time series(TS). 
\subsubsection{Multiple Linear Regression}
Given data $D = (X_{1}, Y_{1}); \cdots; (X_{n}, Y_{n})$ where $X_{i}$ and $Y_{i}$ are the feature vector and response variable, respectively, where $X_{i}= [X_{i1}, X_{i2}, \cdots, X_{ip}]$ represents $p$ variable values. The multiple linear regression (MLR) will take the following form:
\begin{equation}
f(X) = \beta_{0} + \beta_{1}X_{\_1} + \cdots + \beta_{p}X_{\_p} + \epsilon
\label{eq:linear1}
\end{equation}
Where $X_{\_j}$ represents the $j$ th variable and $\beta\_{j}$ quantifies the association between that variable and the response.

\subsubsection{Time Series: ARIMA}
ARIMA stands for Autoregressive Integrated Moving Average models. ARIMA modeling takes into account historical data and decompose it into an autoregressive (AR) process, where there is a memory of past events; and Intergrated (I) process, which accounts for stabilizing or making the data stationary and ergodic, making it easier to forecast; and a Moving Average (MA) of the forecast errors, such that the longer the historical data, the more accurate the forecast will be, as it learns over time. ARMIA models therefore have three model parameters, one for the AR($p$) process, one for the I($d$) process, and one for the MA($q$) process, all combined and interacting among each other and recomposed into ARIMA($p,d,q$) model. In the experiments, the optimal ARIMA model for a given dataset was selected according to Akaike Information Criterion (AIC) \cite{Akaike1973}. That is, AIC was used to determine if a particular model with a specific set of $p, d$ and $q$ parameters was a good statistical fit.

Another potential advantage of time series forecasting is the ability to predict future values based solely on previously observed values. This method will reveal more usage when we move our forecast effort to newly emerging business sector with limited data visibility.

\subsection{Model Selection and Ensemble Process}
In forecast models, the response variables were the weekly CPU billings in each line of business, e.g. Desktop CPU, Notebook CPU and Server CPU. Since the existence of competitor and complimentary relationship among those sectors, for instance, to a certain extent notebook can be considered as substitute product of Desktop, the independent variables were used across line of business. We left around 35 (numerical and categorical) variables for each model which were derived from the categories of variables described in Sec.\ref{sec:data} and past the multicolinearity treatment described in Sec.\ref{sec:colinear}. 

Under the consideration of the ratio between numbers of observations and variables, we faced a problem of performing regression in a high-dimensional space. Sugguested by empirical results, using all the variables would result in over fitting. Traditional approach to tackle this problem was to use step-wise subset selection (forward, backward or hybrid approach) as long as that optimizes the measurement (AIC, BIC, or adjusted $R^2$). One well-known problem from those approaches was that the solutions were not always optimal \cite{Yuan2006,Goodenough2012}. Thanks to the increasing of the computation power and development of parallel execution in data science tools \cite{Weston2015}, we were able to perform best subset selection using brute force technique. The models trained and validated on different subset of variables in parallel. After training, the models were ranked by their performance on validation set. The measurement we used was mean absolute percent errors (MAPE) because of its sensitivity to magnitude changes of the response variable. In validation set MAPE was defined as the average MAPE over all validation samples.

\begin{equation}
\mathrm{MAPE} = \frac{100\%}{m} \sum_{i=1}^{m} \frac{|\hat{Y_i} - Y_i|}{Y_i}
\label{eq:vs1}
\end{equation}

\begin{figure}[ht]
\centering 
   \includegraphics[width=0.75\textwidth]{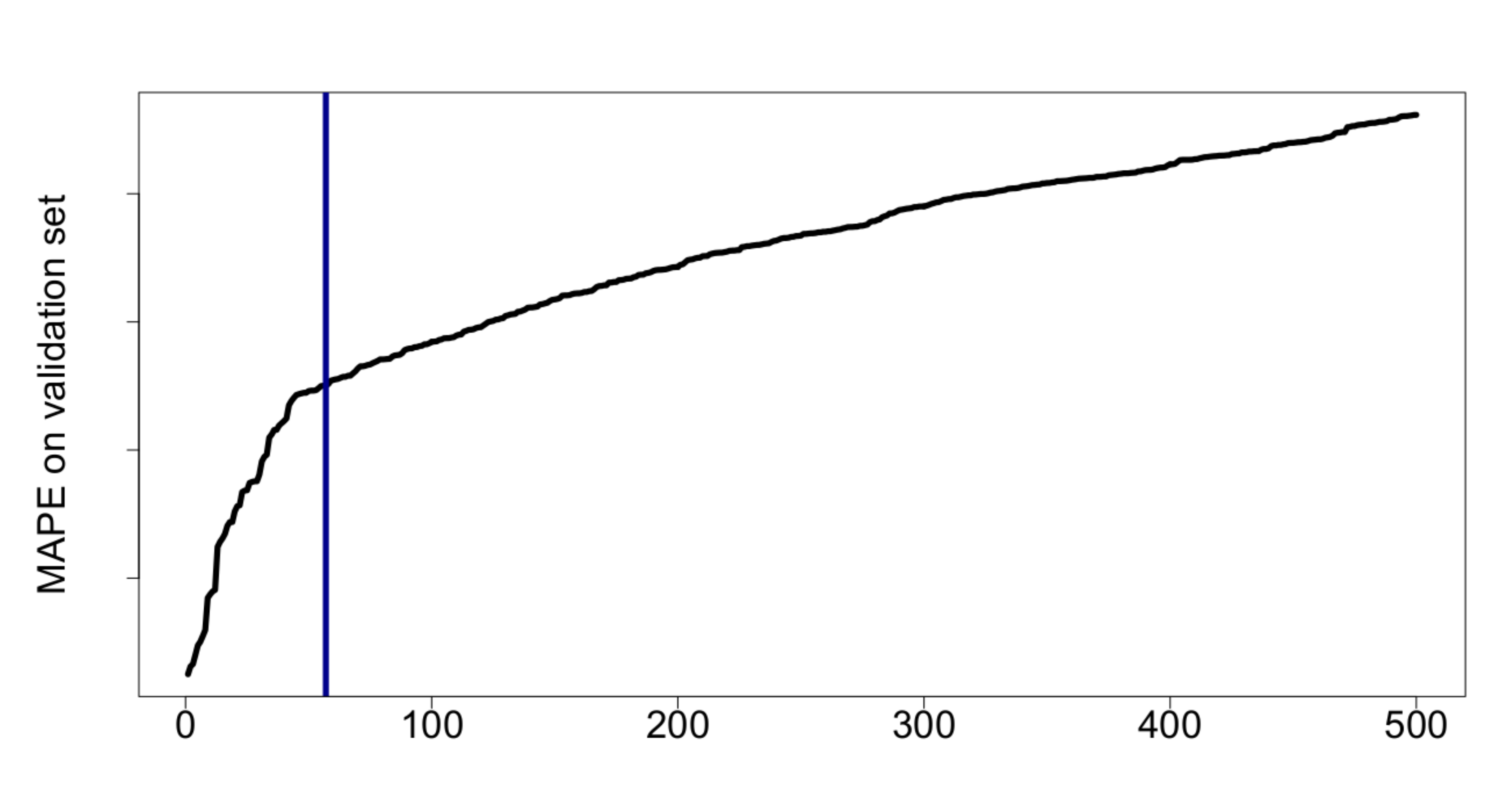}
    \caption{Number of models detected by change point algorithm on MAPE on validation set. X-axis showed the MAPE from each of the model.}
    \label{fig:change_point}
\end{figure}

To take full advantage of the millions of models, we took top $M$ models, instead of a single best model, to form the ensemble model. This number $M$ was determined by a combination of error threshold and change point algorithm proposed in \cite{Killick2014}.  In Fig.\ref{fig:change_point}, the curve shows sorted MAPE on validation set, the vertical line was determined by change point algorithm to show the number of models which performance similar on the validation set and better to be incorporated into the final ensemble stage. In this specific case, the number of models was 57.

Each model among the best $M$ models was denoted by $\hat{f_i}$, the final prediction was given by averaging all the predictions 
\begin{equation}
\hat{f}(X) = \frac{1}{M}\sum_{i=1}^{M}\hat{f}_{i}(X)
\label{eq:vs2}
\end{equation}

This ensemble approach was applied to all methods. Although XGBoost and random forest are already ensemble methods, we still added in another layer of ensemble. From this point on, we will denote the ensemble XGBoost, RF, MLR and ARIMA as EXGBoost, ERF, ELR and ETS respectively. 

\subsection{Moving Window Validation}
\label{sec:window}
Our goal is to forecast weekly CPU sale volume in short term (within 16 weeks). The approach was to build separate model for different lead times. e.g. standing at today time $t$, a model $\hat{f}_{t+1}$ to forecast the next week sale was built independently with the model $\hat{f}_{t+2}$ for forecasting the sale the week after next week. This approach fully considered the unique characteristic of forecasting with various lead times, also led to as large of a reduction in variance as averaging many uncorrelated quantities when we were more interested in monthly or quarterly total. In particular, this approach led to a substantial reduction (in percentage) variance over a multi-week window. 

\begin{figure}[ht]
	\centering 
    \includegraphics[width=0.75\textwidth]{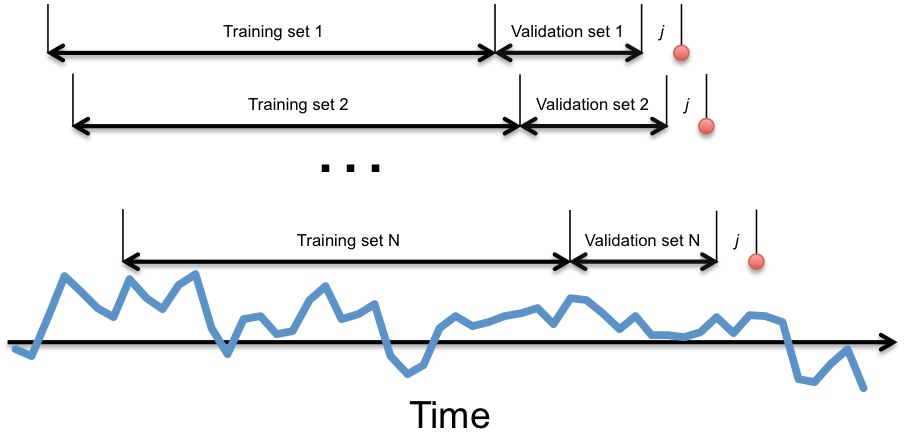}
    \caption{A illustration of training, validation and test(red dots) sets for model $\hat{f}_{t+j}$ with $j$ weeks lead time.}
    \label{fig:validation1}
\end{figure}

For each lead time model $\hat{f}_{t+j}$, we split the total data set into training data (2 years), validate data (1 year) and test data (1 year). For a fix leading time, to test the performance on next observation in the test set, the entire training, validation and test set were moved forward one week and the process was repeated, as depicted in Fig.\ref{fig:validation1}. This moving window framework considered the time-dependence of the data and would automatically select the models which represented back-then up to date market condition.

\subsection{Variable Importance}
\label{sec:variable}
Ensemble models typically result in improving accuracy and reducing variance versus a single model. However, Ensemble models are more difficult to interpret. In practical application, the lack of interpretability made the model adoption extremely hard. To overcome this, we borrowed of idea of permutation test and proposed a generic method to obtain an importance summary of variables.

The method was summarized as follow: For the $M$ models $\hat{f}_i, i = 1, \cdots, M$ selected into the ensemble process, we recorded the $\mathrm{MAPE}_0$ defined as in \ref{eq:vs1} of the validation set. 
For each variable $X_{\_j}$, repeat the following steps for number of times.
\begin{enumerate}
\item Permute or ``shuffle" the value of $X_{\_j}$ over all the data (e.g., by assigning different values to each observation from the set of actually observed values) without replacement. 
\item Repeat the model training on the training set
\item Generate forecast result for observation in validation set
\item Calculate the $\mathrm{MAPE}_j$ defined as \ref{eq:vs1} on the validation set and record $\mathrm{MAPE}_j$.
\end{enumerate}

Over a sufficient amount of iterations (we used 100, although empirical result suggested results stabilized after 60 - 70 iterations.), the average $\mathrm{MAPE}_j$ was calculated and denoted as $\overline{\mathrm{MAPE}}_j$. The difference between $\overline{\mathrm{MAPE}}_j - \mathrm{MAPE}_0$ was used as the importance measurement of variable $j$,  while a large value of  $\overline{\mathrm{MAPE}}_j - \mathrm{MAPE}_0$ indicated a more important variable.

Since the variable importance analysis only needed to run on $M$ models, the computation was efficient. Without assuming certain distribution, this non-parametric permutation variable importance analysis provided a general framework across methods to evaluate the variable importance and to increase model interpretability. This approach could also be extended to classification problem by modifying the MAPE into measurement of misclassify error.

\section{Empirical Evaluation}
\label{sec:result}
We would dive deep into the performance comparison of ELR, ERF, ETS and EXBoost for lead time (1-16 weeks) in this section. For various lead times, we fixed the test set as the weekly CPU sale of 2015 (Intel Calendar). The training data started from 2012. The training, validating and testing followed the moving windows described in Fig.\ref{fig:validation1}.

Models were tested on 52 target weeks in 2015 with 16 different lead times for different lines of business. The performance of those models was compared at the target week and lead time level. The detailed ranking counts were shown in Table \ref{tab:perf_DT}. Besides the performance of 1 - 16 weeks of lead time, we also reported the model performance of shorter lead time (1 - 5) and longer lead time (12 - 16). The three numbers in each of the table show the counts for lead time 1-16, 1-5 and 12-16, respectively. For instance, we applied all four methods to forecast 1st week's sale with 1 week lead time, then the performance of those four methods were ranked by comparing MAPE on test samples. The ranking of each method was accumulated for the test weeks and the lead times. In Table \ref{tab:perf_DT}, take first row of desktop performance as an example, out of 832 ($=$ 52 forecast weeks $\times$ 16 lead time weeks) ELR experiments, 185 performed the best, 199 ranked 2nd, 221 ranked third, and 227 showed the worst results. The rank counts in the parentheses gave us a more detailed view of short-term ($\leq5$ weeks) and long-term ($\geq12$ weeks) performance. Among 260 (52$\times$5) ELR short-term experiments for desktop, 58 achieved the best results, while 58, 71 and 73 ranked 2nd - 4th place respectively. The following conclusion could be drawn from this table: for different lines of business, the forecasting performance generated by different forecasting models were mixed, for instance, ERF performed better for desktop CPU forecasting, ELR generated better results for server, and EXBoost showed bipolar performance in notebook and server; another observation was that even for the same model, the lead time could have large effect on forecasting results, for instance, ETS performed better for shorter lead time forecasting on desktop.

\begin{table*}[ht]
\centering
 \begin{tabular}{|| c | c | c | c | c ||} 
 \hline
 Performance Ranking for DT & 1 & 2 & 3 & 4 \\ 
 \hline
 ELR & 185 (58, 65) & 199 (58, 86) & 221 (71, 55) & 227 (73, 54) \\ 
 \hline
 ERF & 231 (73, 64) & 206 (72, 54) & 211 (59, 77)& 184 (56, 65)\\
 \hline
 ETS & 216 (75, 56) & 206 (60, 61) & 218 (62, 71) & 192 (63, 72) \\
 \hline
 EXBoost & 200 (54, 75) & 221 (70, 59) & 182 (68, 57) & 229 (68, 69) \\ 
 \hline\hline
 
 Performance Ranking for MB & 1 & 2 & 3 & 4\\ 
 \hline
 ELR & 211 (73, 63) & 201 (66, 69) & 212 (69, 49)& 208 (52, 79) \\ 
 \hline
 ERF & 174 (49, 61) & 259 (75, 78) & 227 (72, 68)& 172 (64, 53)\\
 \hline
 ETS & 201 (59, 63) & 218 (61, 69) & 204 (63, 67) & 209 (77, 61) \\
 \hline
 EXBoost & 246 (79, 73) & 154 (58, 44) & 189 (56, 76) & 243 (67, 67) \\ 
 \hline\hline
 
 Performance Ranking for SVR & 1 & 2 & 3 & 4\\ 
 \hline
 ELR & 227 (63, 72) & 232 (71, 70) & 203 (72, 60)& 170 (54, 58) \\ 
 \hline
 ERF & 177 (57, 50) & 221 (75, 65) & 233 (68, 77)& 201 (60, 68)\\
 \hline
 ETS & 196 (58, 73) & 229 (63, 76) & 221 (75, 65) & 186 (64, 46) \\
 \hline
 EXBoost & 232 (82, 65) & 150 (51, 49) & 175 (45, 58) & 275 (82, 88) \\ 
 \hline
\end{tabular}
\caption{Performance ranking for models on weekly sale prediction for Desktop, Notebook and Server. This table showed the counts of each method for target weeks and lead times. The three numbers in each cell represented counts for all lead times, counts for lead time $\leq$ 5 and counts for lead time $\geq12$ .}
\label{tab:perf_DT}
\end{table*}

\begin{figure}[ht]
\begin{subfigure}{0.32\textwidth}
    \includegraphics[width=1\textwidth]{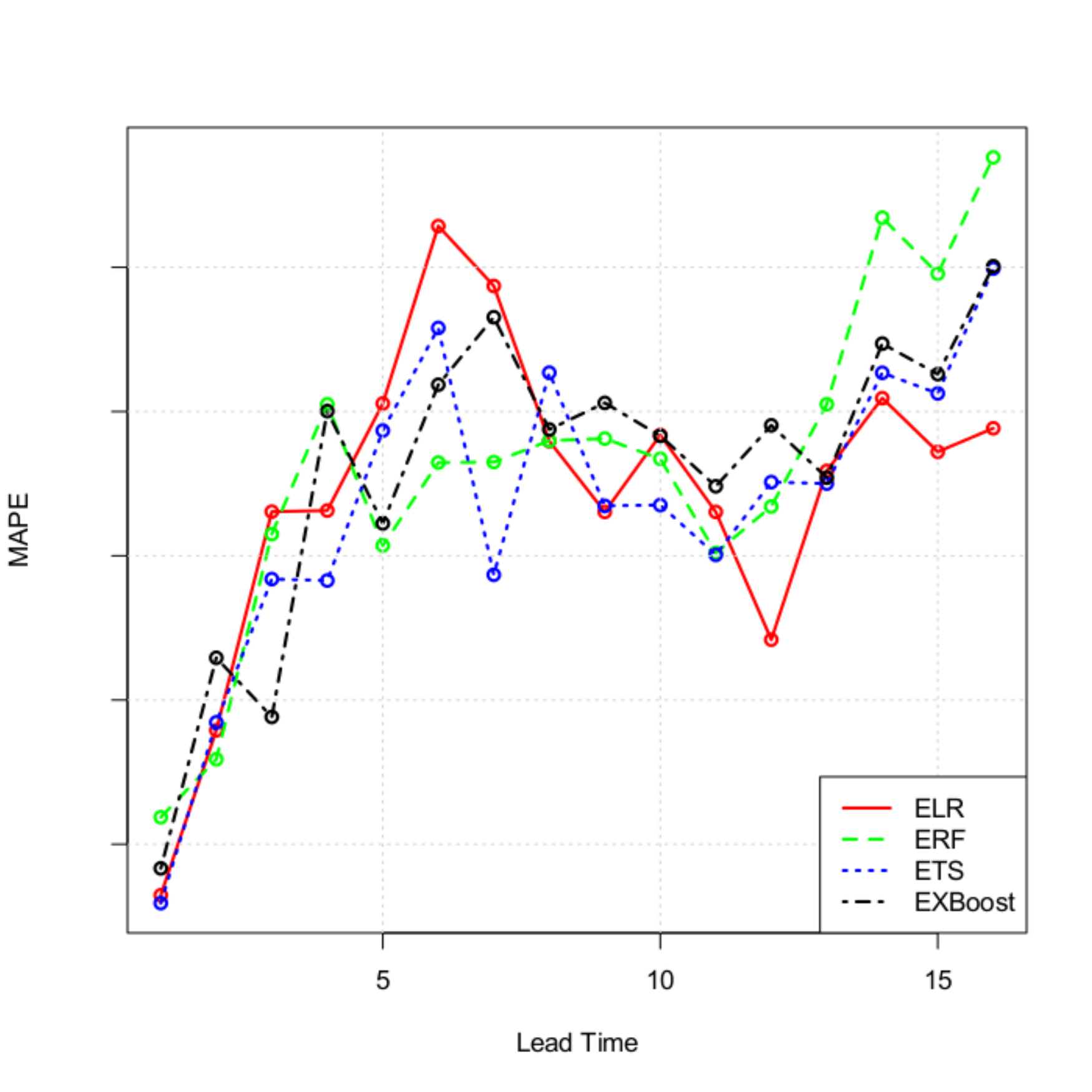}
    \caption{}
    \label{fig:DT_lead}
\end{subfigure}
~
\begin{subfigure}{0.32\textwidth}
    \includegraphics[width=1\textwidth]{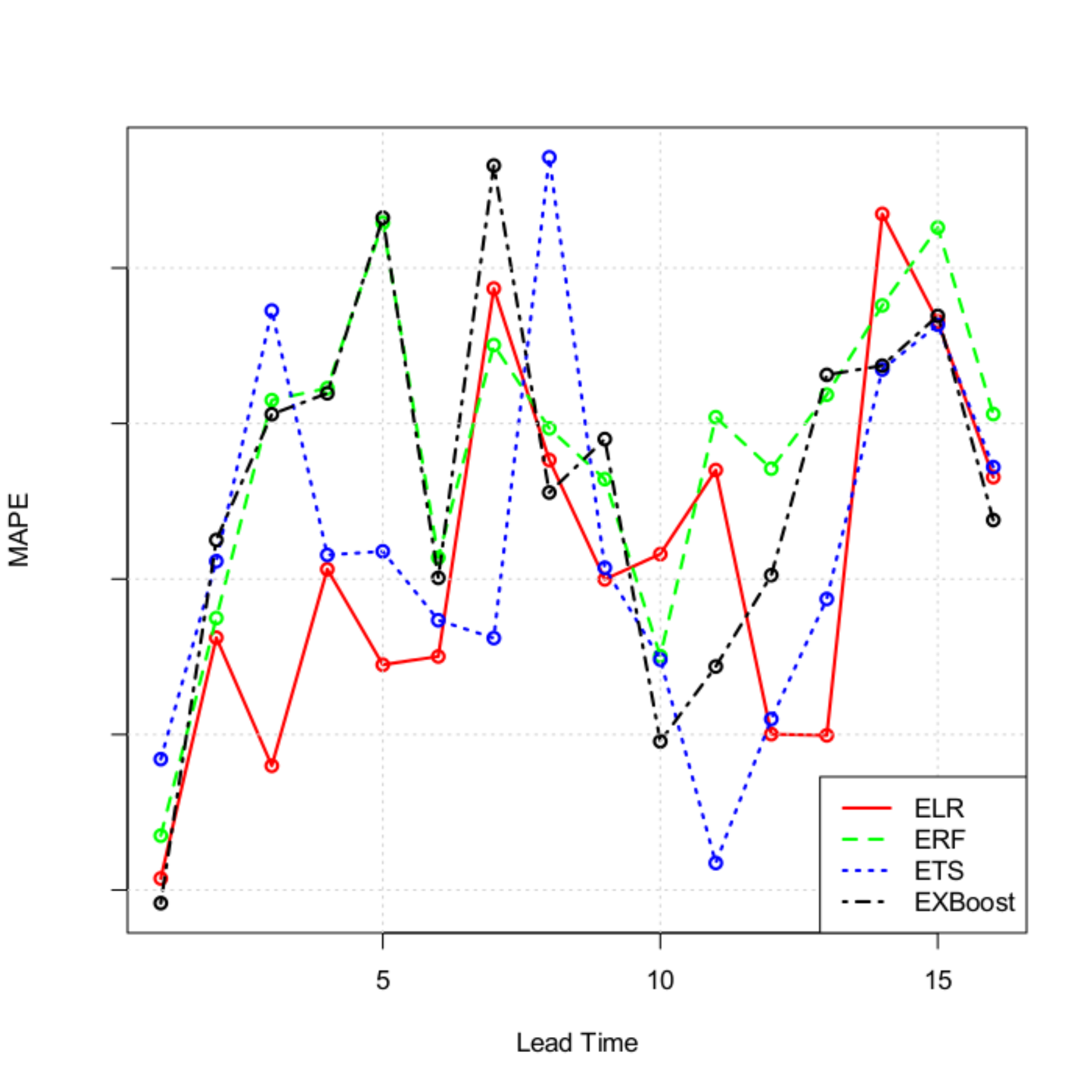}
    \caption{}
    \label{fig:MB_lead}
\end{subfigure}
~  
\begin{subfigure}{0.32\textwidth}
    \includegraphics[width=1\textwidth]{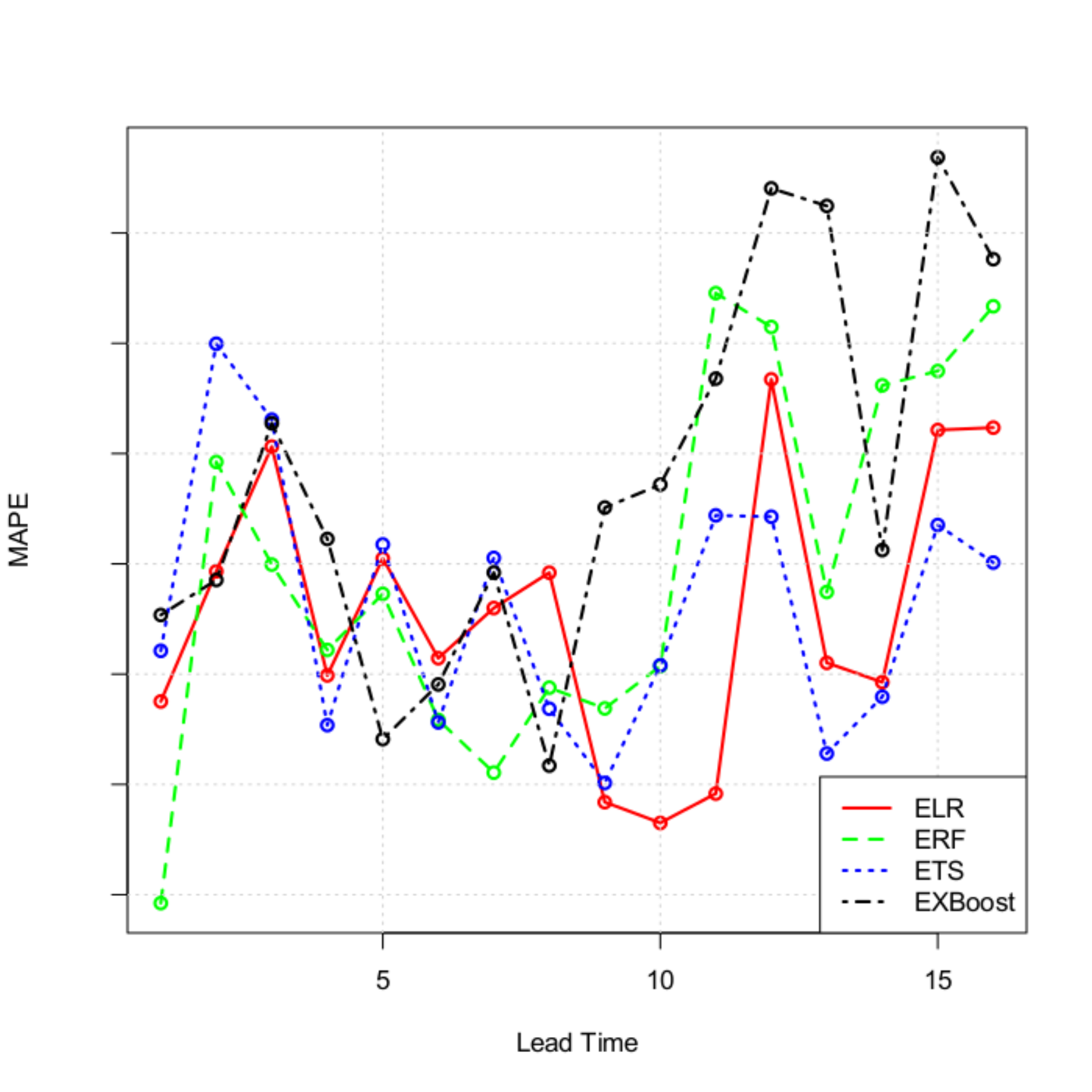}
    \caption{}
    \label{fig:SVR_lead}
\end{subfigure}
\caption{Average MAPE on test set of models along lead time(in weeks) on Desktop(a), Notebook(b) and Server(c).}
    \label{fig:CPU_lead}
\end{figure}

To better understand the model performance on various lead times, we calculated the average MAPE across all the test weeks by lead time. Figure \ref{fig:CPU_lead} showed the change of MAPE along lead time. Surprisingly, the MAPE did not show monotonically increase as lead time increased, although lead time of 1 week showed the lowest MAPE. This indicated the high dynamic nature of the market. Insights could be drawn from those figures, for instance, in desktop forecasting shown in Fig.\ref{fig:DT_lead}, ELR showed better results than other methods in longer lead time forecast; in server forecast as in Fig.\ref{fig:SVR_lead}, combination of ERF for shorter forecasting and EXBoost for longer forecasting would outperform other methods.

\begin{figure}[ht]
\centering
\begin{subfigure}{0.4\textwidth}
 \centering
    \includegraphics[width=1\textwidth]{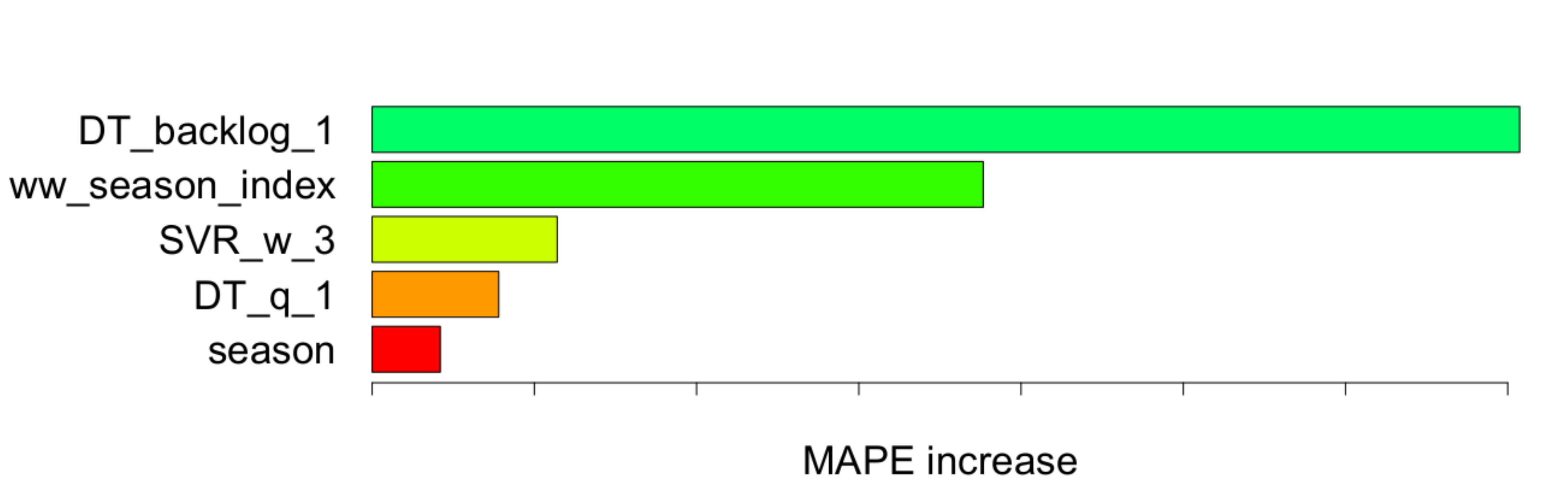}
    \caption{ELR}
    \label{fig:LR_importance}
\end{subfigure}   
~
\begin{subfigure}{0.4\textwidth}
 \centering
    \includegraphics[width=1\textwidth]{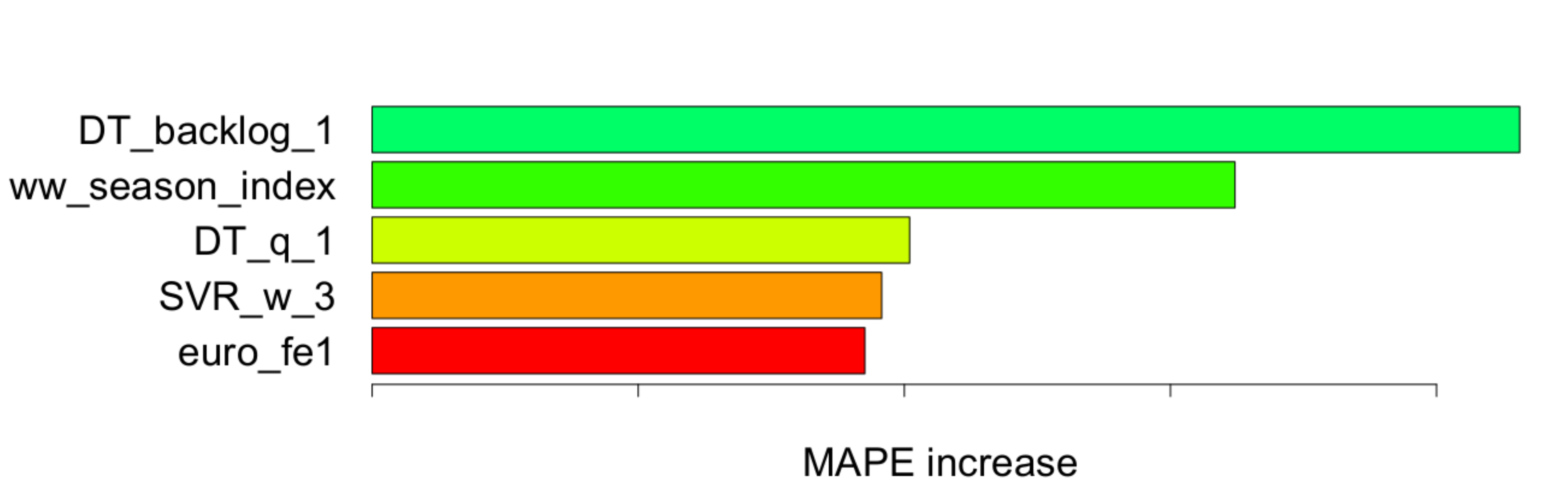}
    \caption{ERF}
    \label{fig:RF_importance}
\end{subfigure}

\begin{subfigure}{0.4\textwidth}
 \centering
    \includegraphics[width=1\textwidth]{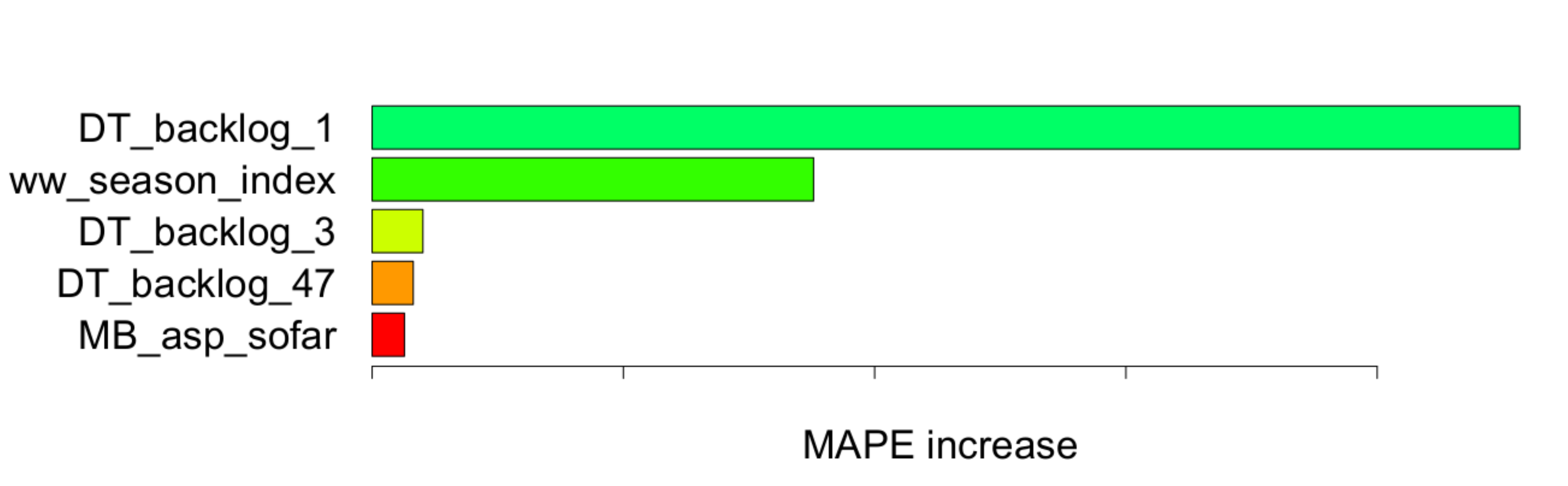}
    \caption{ETS}
    \label{fig:TS_importance}
\end{subfigure} 
~
\begin{subfigure}{0.4\textwidth}
 \centering
    \includegraphics[width=1\textwidth]{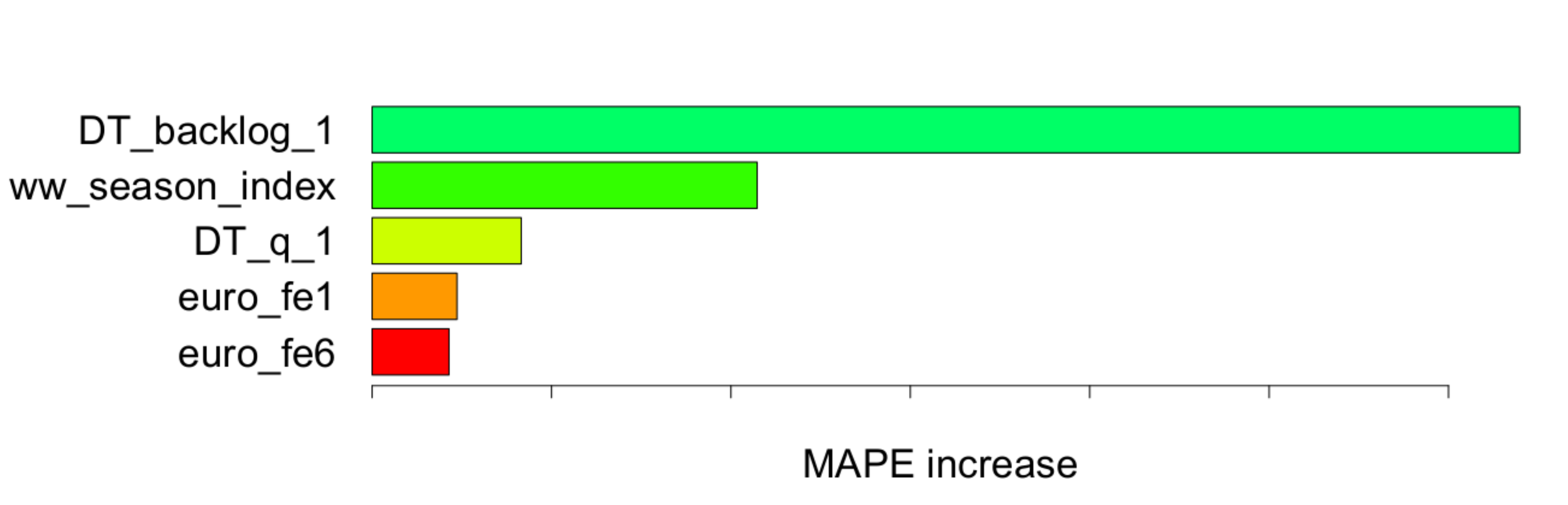}
    \caption{EXBoost}
    \label{fig:Boost_importance}
\end{subfigure}
\caption{Top 5 important variables of models for a Desktop example}
\label{fig:importance}
\end{figure}

As we proposed in Sec.\ref{sec:variable}, the variable importance was examined for each ensemble model. To show a clearer comparison among those methods, we fix the target week (1st week of 2015), lead time (1 week) and line of business (DT). The top 5 most important variables for each method were shown in Fig.\ref{fig:importance}. In those figures, the x-axis showed the MAPE increase on the validation set by permuting the value of certain variable. All methods identified DT$\_$backlog$\_$1 \footnote{The booking amount for back then next week} and ww$\_$season$\_$index \footnote{Week index in a quarter} as the top two important variables, with the DT$\_$backlog$\_$1 as the dominant one. This showed that in the shorter term, the forward booking had strong predictive power, while the constant appearance of ww$\_$season$\_$index\ indicated the weekly sale had strong in-quarter-selling-pattern. For this specific example, the effect of top 2 variables was more dominant in ELR, ETS and EXBoost, while the effect of top variables in ERF more spread out. We also noticed the appearance of cross sectors variables, which implied the success of using cross sectors variables. For example, the SVR$\_$w$\_$3 \footnote{The Server CPU sale 3 weeks ago} and MB$\_$asp$\_$sofar \footnote{The average selling price of notebook} showed up as important variables in this desktop forecast example.

Comparing all the empirical results from various methods, the following conclusion can be drawn:
\begin{enumerate}
\item For different lines of business, the forecasting performance generated by different forecasting models were mixed. For current quarter performance tracking (the total forecast for the weeks left in this quarter) and next quarter guidance generating (the total forecast for the weeks in next quarter), our current approach was to take the average output of all the models. Back track test results and live testing indicated that this approach was superior with less effort than the traditional bottom-up approach.
\item For different lead times, the forecasting error was not monotonically increasing as lead time increased.
\item For different lines of business and lead times, the combination of various models might yield better granular results. For example, averaging model outputs with inverse MAPE weighting is one possible approach. We will further investigate the performance of this approach.
\item Important variable analysis suggested strong correlation between certain variables and response. However, this correlation effect did not necessarily lead to concrete conclusion regarding the existence or the direction of causality relationship.
\end{enumerate}

\section{Significance and Impact}
\label{sec:conclusion}
In this paper, we shared our work on the development, implementation and comparison of a sale forecasting tool to predict Intel's weekly CPU sale by lines of business. The novel feature engineering helped to  reduce the subjective bias and measurement errors, especially for economic indicators. The time sensitive training, validating and testing framework allowed the model to effectively reflect changes in the environment and to continuously be evaluated and revised to maintain its credibility. The comparison among those models showed mixed results. This suggests that no general conclusion drawn as to which is the best forecasting technique to employ, but it is certainly true that a forecast should consider multiple models by taking into account the available data resources, characteristic of business and the requirement of the forecast. Our variable importance algorithm boosted the interpretability of the ensemble models and enabled stakeholders examining and giving timely feedback. Those positive aspects of our model led to the recent model adoption. Further work will be 1) To break the forecasting problem into finer granularity and to forecast at geography level and customer level; 2) To develop hybrid techniques. The current approach considers each model individually. In order to break the model boundaries, we plan to take the multi-stage approach. One model can serve as the pretreatment of the data or the feature engineering of variables, thus, to prepare better features for other models. This is similar to the idea of deep belief networks; 3) To combine the intelligence of human and the computing power of machine to capture other (can not observe in data) factors.

\section{Acknowledgment}
We want to thank Aziz Safa, Vice President of Information Technology, Mary Loomas, the Sr. Controller of World Wide Revenue and Aaron Smith, the manager of World Wide Revenue for their continuing support and sharing valuable business expertise through discussions. 
We also thank numerous other executives and employees for their assistance and support throughout this project. 

\bibliography{ICDM}
\bibliographystyle{abbrv} 
\end{document}